# Identification of Play Styles in Universal Fighting Engine

Kaori Yuda, Shota Kamei, Riku Tanji, Ryoya Ito, Ippo Wakana and Maxim Mozgovoy
The University of Aizu
Tsuruga, Ikki-machi, Aizu-Wakamatsu, Fukushima, 965-8580 Japan
E-mail: {m5241108, s1270112, s1270139, s1260233, s1260210, mozgovoy}@u-aizu.ac.jp

**KEYWORDS**
Turing test, behavior analysis, fighting game

**ABSTRACT**
AI-controlled characters in fighting games are expected to possess reasonably high skills and behave in a believable, human-like manner, exhibiting a diversity of play styles and strategies. Thus, the development of fighting game AI requires the ability to evaluate these properties. For instance, it should be possible to ensure that the characters created are believable and diverse. In this paper, we show how an automated procedure can be used to compare play styles of individual AI- and human-controlled characters, and to assess human-likeness and diversity of game participants.

**INTRODUCTION**
Fighting games provide a variety of interesting challenges for AI research and development. Fighting is often seen as a purely arcade fun, emphasizing fast reaction and the ability to perform complex combo actions with accurate timing. However, numerous current research works show that designing a good AI for a fighting game is not an easy task.

First, achieving high performance is challenging: only one recent work (Oh et al. 2021) reports obtained AI skill level comparable to the abilities of professional human players in a modern fighting game. One of the principal difficulties lies in the fact that people both *react to* and *anticipate* opponent's movements. Thus, a fighting game can be considered as a rock-paper-scissors type game, where opponents make "double-blind decisions" (Yu and Sturtevant 2019). Second, people of different skill levels need AI opponents possessing comparable and possibly adjustable skills, which is a separate challenge (Ishihara et al. 2018). Finally, AI-controlled characters have to be *believable* (human-like) and exhibit *diverse play styles* to keep the players engaged. Believability and diversity of AI behavior is not a universal requirement across game genres, but for certain game types, such as first-person shooters, it seems to be the case (Soni and Hingston 2008). Fighting games typically simulate a one-vs-one combat between two human-like opponents, so certain "human-like traits" are expected form the AI system, at least as a feature contributing to the "realism" of the environment.

Before engaging in the task of creating believable and diverse AI-controlled characters for a fighting game, one has first to confirm that the game environment used is able to provide sufficient flexibility for this work. In other words, it should be possible for game characters to exhibit diverse play styles, recognizable by human observers and identifiable distinguishable with a certain evaluation method.

The goal of this paper is to analyze play styles of human- and AI-controlled characters in a Universal Fighting Engine (UFE) (Mind Studios 2021). We develop a simple procedure, able to distinguish individual players, which supports the presumption that identifiable behaviors are achievable in UFE. We also compare play styles of people with the style exhibited by a built-in AI system. Finally, we report results of a short survey, aimed to reveal whether human observers can spot "human-like" traits in the behavior of game characters.

**UNIVERSAL FIGHTING ENGINE PLATFORM**
Universal Fighting Engine (Mind Studios 2021) is a highly customizable platform for one-vs-one fighting games developed in Unity. It supports a large variety of attacks and special moves as well as the ability to create new action types on per-character basis. UFE aims to provide classic combo-heavy 2D fighting gameplay, associated with games such as Street Fighter or Mortal Kombat.

UFE comes with a built-in customizable rule-based AI engine called "Fuzzy AI". It relies on fuzzy logic to evaluate the current scene and estimate the desirability of each given action. Play style and skill level of Fuzzy AI players can be adjusted by tuning a number of parameters, by default organized into presets ranging from "very easy" to "impossible". In the present study we use five different levels of Fuzzy AI with default parameter values shown in Table 1.

Table 1: Fuzzy AI settings for five different skill levels

|  | Very easy | Easy | Normal | Hard | Very hard |
|---|---|---|---|---|---|
| Time between decisions | 0.4 | 0.3 | 0 | 0.1 | 0 |
| Time between actions | 0.1 | 0.1 | 0.05 | 0.05 | 0.05 |
| Rule compliance | 0.9 | 0.9 | 0.9 | 0.9 | 0.9 |
| Aggressiveness | 0.1 | 0.3 | 0.5 | 0.6 | 0.6 |
| Combo efficiency | 0.1 | 0.2 | 1 | 1 | 1 |

For us, each AI preset is essentially a "black box" aimed to represent a unique fighting game character. Thus, we will not discuss the choice of Fuzzy AI parameters and their values, described in the documentation as follows:
- **Time between decisions (sec)**: minimum time taken to formulate a decision.
- **Time between actions (sec)**: time between executing each decision.
- **Rule compliance**: controls the balance between systematic appliance of rules and randomicity (higher values correspond to lower randomicity).
- **Aggressiveness**: controls the balance between basic moves such as walk, crouch and jump, and attacks. Higher values correspond to higher contribution of attacking actions.

- **Combo efficiency**: controls the probability of attempting combo actions.

Universal Fighting Engine comes with a set of pre-modeled characters, distinct in their special move types. To ensure fair comparison, we use the same character type for each of the opponents in all test games.

## PLAY STYLE SIMILARITY IDENTIFICATION

The goals of our work can be narrowed down to the following research questions:

RQ1  Do human-controlled and AI-controlled characters possess distinct, identifiable play styles?
RQ2  Are these styles consistent across matches or change depending on the opponent?
RQ3  Do human-controlled characters possess identifiable "human-like behavior traits"?
RQ4  Can questions RQ1-RQ3 be answered with a certain automated evaluation procedure?

In order to compare individual players' behavior, we adopted a cosine similarity-based procedure, earlier used in the game of boxing (Mozgovoy and Umarov 2010). It operates as follows. We analyze recordings of games where a character of our interest participates and create its "behavior fingerprint" as an ordered list of probabilities of every possible tuple ($A_1$, $A_2$, $A_3$), representing three consecutive player actions. Recordings consist of game engine state snapshots taken at each consecutive simulation frame. Within this context, each action is uniquely defined with its game engine-specified elements `currentState`, `currentSubstate`, and `currentBasicMove`. While more details are provided in Table 2, we view these elements as merely items uniquely identifying an action in our particular game engine (UFE).

Being lists of probabilities, behavior fingerprints can be compared as vectors using cosine similarity, yielding a similarity ratio of [0, 1]:

$$Similarity = \frac{A \cdot B}{\|A\|\|B\|}$$

Since two players participate in each recording, it is possible to compare fingerprints of the same player obtained in matches with different opponents.

Table 2: Elements of player actions

| | |
|---|---|
| `currentState` | Indicates character's state (e.g., Stand, Jump, Down) |
| `currentSubState` | Indicates additional state modifier of the character (e.g., Resting, Blocking) |
| `currentBasicMove` | Indicates character's basic movement state (e.g., Idle, MoveForward) |

Behavior fingerprint comparison was applied to a dataset consisting of matches, played by four humans and five skill presets of Fuzzy AI as follows:

1) Every human participant played with every other human participant and with an AI system set to a normal skill level (10 player pairs).
2) Every AI-controlled character played against another AI-controlled character, set to a different skill level (10 player pairs).
3) Two human participants played against each AI skill preset (10 player pairs).

Every pair played 10 matches of two rounds. A round lasts 100 seconds unless a knockout occurs. A separate behavior profile for a particular player is built by processing all 10 matches played against a specific opponent.

**Similarity Identification Results**

Our first observation concerns play style diversity of Fuzzy AI. Any two given behavior profiles of AI-controlled characters have a similarity ratio of at least 76%, and most of the scores are higher (88% on average). This way, we can conclude that parameter tuning has very little effect on AI behavior profiles. While the skill level of the AI system can be modified, its play style remains virtually the same.

Thus, for the sake of simplicity, we will mostly deal with AI set to the "normal" difficulty in subsequent tests. It should be noted, however, that high play style similarity values were obtained with an automated scoring algorithm, it may not fully agree with human perception of a play style.

Interestingly, human players seem to adhere to the same play style even when they face different opponents. The similarity ratio between two behavior profiles of the same player obtained in matches against a variety of opponents is at least 80% on average (see Table 3). Lower similarity scores were obtained in matches against "very easy" and "very hard" AI-controlled characters. Thus, people seem to be more inclined to modify their play style as a response to different degrees of challenge rather to different play styles of their opponents.

Table 3: Similarity scores for profiles
of the same character in different matches

| | Minimum | Maximum | Average |
|---|---|---|---|
| AI-normal | 0.76 | 0.98 | 0.88 |
| Ippo | 0.61 | 0.93 | 0.82 |
| Kaori | 0.70 | 0.94 | 0.85 |
| Ryoya | 0.69 | 0.99 | 0.85 |
| Riku | 0.70 | 0.88 | 0.80 |

The observations above greatly simplify subsequent comparisons of player profiles. Instead of dealing with context-sensitive profiles (such as "player A as seen in games against player B"), we can discuss generalized profiles of individual players, representing their typical play style across game sessions. The results of a direct comparison of these profiles using cosine similarity function are summarized in Figure 1 and Table 4.

Our calculations show that people are indeed closer to other people in terms of their play style. Only one participant (Ryoya) happened to be closer to AI than to other people. It is also clear that individual play styles are distinguishable: as mentioned before, comparison of behavior profiles of the same player obtained in matches with different opponents typically yields values of 90% and above, while the average similarity between different players is only 44%-65%.

Table 4: Play style similarity between human-controlled and AI-controlled characters

|  | Similarity with AI | Average similarity with human players | Median similarity with human players |
|---|---|---|---|
| AI | 1.0 | 0.46 | 0.46 |
| Ippo | 0.54 | 0.65 | 0.68 |
| Kaori | 0.38 | 0.53 | 0.62 |
| Ryoya | 0.73 | 0.58 | 0.62 |
| Riku | 0.18 | 0.44 | 0.44 |

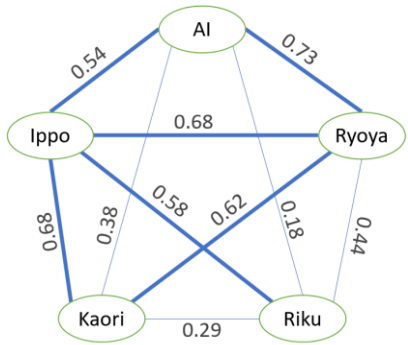

Figure 1: Play style similarity of game participants

**ASSESSING BELIEVABILITY WITH TURING TEST**

While higher similarity scores for human participants suggest the existence of identifiable "human-like traits" in their play styles, an automated procedure alone cannot serve as a reliable proof of this suggestion. Livingstone (2006) discusses the possibility to apply a variation of Turing test (Turing 1950) to evaluate human-likeness of computer-controlled characters. In this scenario, believability is judged by people rather than by an automated scoring procedure. Implementing Turing test for a computer game is not a straightforward process, but certain specific recommendations have been proposed in literature (Gorman et al. 2006; Hingston 2009).

To verify conclusions suggested by our play style similarity evaluation method, we performed a quick Turing test-inspired survey, designed to reveal whether external observers are able to distinguish human-controlled and AI-controlled characters in UFE. The survey was conducted online using Google forms. The participants were asked to watch four video clips (ranging from 1m06s to 3m15s in length) of matches between two unknown participants. The task was to guess which characters are controlled by people, and which by the AI system. To reduce random guessing, a third "not sure" choice was also available. In total, 14 subjects participated in the survey. One of them is over 40 years old, and the rest are 20-25 years old. Most of the participants (71%) identified themselves as occasional players of fighting games, while the rest pointed that they do not play fighting games at all.

**Survey Results**

Table 3 shows aggregated survey results for each video clip. While our small sample size does not allow to make reliable conclusions, certain observations can still be made.

Clearly, distinguishing human- and computer-controlled opponents in a fighting game is not easy: our participants were wrong more often than they were right. The results are slightly better for occasional fighting game players, who provided an equal number of correct and incorrect answers (45%), choosing "not sure" option only in 10% of cases. It might be even hard to judge consistently: both participants in Match 3 are identical AI-controlled characters, but for some reason our observers found more human-like traits in the left-hand side character. Curiously, the most "AI-like" player Ryoya (according to the cosine similarity method), also got the lowest number of correct answers in the Turing test.

Table 3: Turing test results

|  | Match participants | Correct answers | Incorrect answers | "Not sure" answers |
|---|---|---|---|---|
| Match 1 | P1: Riku | 43% | 57% | 0% |
|  | P2: Ryoya | 36% | 57% | 7% |
| Match 2 | P1: Kaori | 50% | 21% | 29% |
|  | P2: Ryoya | 29% | 36% | 36% |
| Match 3 | P1: AI | 43% | 36% | 21% |
|  | P2: AI | 29% | 43% | 29% |
| Match 4 | P1: Kaori | 36% | 36% | 29% |
|  | P2: AI | 29% | 43% | 29% |
| **Total** |  | **37%** | **41%** | **22%** |

As a follow-up to this test, we had a discussion of five clips (two human-human, two human-AI and one AI-AI game) with three experienced gamers who play fighting games at least weekly for over than 3 years. One person could correctly identify whether a certain player is human or AI in 8 cases out of 10, another individual correctly identified 6 players, and the third one made only two correct guesses.

Interestingly, these experts had different ideas about what constitutes "human-like" style. For example, the best-performing individual associated AI behavior with "nice" smooth movements. Another correct guess was made by associating "seemingly intentional" jumps over the opponent with human-like behavior. These remarks somewhat overlap with the comments provided in (Gorman et al. 2006), revealing that relatively simple clues are often associated with human-like or AI-like behavior ("fires for no reason, must be human", "stand and wait, AI wouldn't do this", etc.) As a side note we can add that the *development* of human-like AI is usually done through a variety of "ghosting" or "mirroring" strategies that aim to reproduce actual patterns of human behavior rather than to understand and implement specific features considered "human-like" by other players (Polceanu 2013; Mozgovoy et al. 2016; Schrum et al. 2010).

**DISCUSSION AND CONCLUSION**

The results obtained in the course of the present study suggest the following answers to our research questions:

RQ1. Yes, game characters possess distinct and identifiable play styles in a sense that it is possible to cluster players according to their play styles and find out whether a particular behavior profile belongs to a specific player.
RQ2. Yes, individual play styles are consistent and recognizable even in games against different opponents.
RQ3. "Human-like behavior traits" are seemingly possible to detect with a cosine similarity-based tool described above. At least, this is true for Fuzzy AI system of UFE, but the answer might be different for other AI engines.
RQ4. Yes, an automated approach can be used to address the challenges listed in RQ1-RQ3.

Our behavior comparison tool based on cosine similarity provided consistent and reliable results in most cases. We have to note though that its "similarity scores" should not be taken at face value. The tool only captures certain basic behavior traits; thus, its output allows us to make statements like "Kaori's play style is closer to Ippo's rather than Riku's", but any numerical evaluations are rough.

The results of Turing test are harder to interpret. First, let us note that it is difficult for people to assess numerous video clips and compare play styles of players acting in non-adjacent game sessions. Thus, we have to limit surveys to a small number of short clips. Next, it seems that the ability to assess a play style improves with experience, but even hardcore fighting game players have difficulties in distinguishing AI-controlled characters from human players. Finally, high "human-likeness" scores of the AI system may indicate that Fuzzy AI is indeed a high-quality system, able to imitate patterns characteristic for human players. It may also show that believability of behavior cannot be assessed within few short videoclips, and deeper immersion into the game world is necessary.

We started with a suggestion that a successful fighting game environment should be sufficiently sophisticated to let the players exhibit diverse, identifiable play styles. A good AI system, in turn, should be able to utilize these capabilities, facilitating the development of diverse AI-controlled characters. Players and observers do not necessarily formulate their impression in terms of "human-likeness" or "diversity", but they usually can tell which of the given game worlds is more fun and immersive. Thus, we believe that human evaluation should be used to understand the overall quality of the game, while smaller-scale details such as "behavior similarity of players A and B" can be assessed with a certain automated procedure.